\documentclass{article} 
\usepackage[preprint]{colm2026_conference}

\usepackage{microtype}
\usepackage{hyperref}
\usepackage{url}
\usepackage{booktabs}

\usepackage{lineno}

\definecolor{darkblue}{rgb}{0, 0, 0.5}
\hypersetup{colorlinks=true, citecolor=darkblue, linkcolor=darkblue, urlcolor=darkblue}

\usepackage{amsmath,amssymb,amsfonts}
\usepackage{multirow}
\usepackage{array}
\usepackage{enumitem}
\usepackage{xcolor}
\usepackage{graphicx}
\usepackage{float}
\usepackage{longtable}
\usepackage{tabularx}
\usepackage{titlesec}
\usepackage{caption}
\usepackage{setspace}

\hypersetup{
    colorlinks=true,
    linkcolor=blue,
    citecolor=blue,
    urlcolor=blue
}

\title{UnityMAS-O: A General RL Optimization Framework for LLM-Based Multi-Agent Systems}

\author{%
\makebox[\textwidth][c]{%
\begin{tabular}{c}
\textbf{Renmin University of China \quad Xiaohongshu Inc.}\\[-1pt]
\small Code: \url{https://github.com/chenyiqun/UnityMAS-O}\\[-1pt]
\small See \hyperref[sec:contributions]{Contributions section} for the full author list.\thanks{Project lead and primary contributor: Yiqun Chen. Corresponding author: Jiaxin Mao. Work conducted during Yiqun Chen's internship at Xiaohongshu Inc. Contact: \texttt{chenyiqun990321@ruc.edu.cn}, \texttt{maojiaxin@gmail.com}.}
\end{tabular}%
}%
}

\begin{document}

\ifcolmsubmission
\linenumbers
\fi

\maketitle

\begin{abstract}
LLM-based multi-agent systems decompose complex tasks into interacting roles, but most existing systems remain manually orchestrated: workflows are specified by prompts, tools, and control rules, while the agents themselves are rarely optimized through a unified reinforcement learning interface. Meanwhile, modern RL post-training frameworks mainly target single-policy optimization and do not directly expose the abstractions needed to optimize arbitrary user-defined multi-agent systems with structured interaction, role-specific credit assignment, and configurable parameter sharing.

We present \textbf{UnityMAS-O}, a general RL optimization framework for LLM-based multi-agent systems. UnityMAS-O treats a complete multi-agent workflow as the unit of optimization, rather than a single response or a single policy trajectory. It represents arbitrary task workflows through four first-class objects: logical agent roles, graph-structured trajectories, user-defined reward functions, and explicit agent--model mappings. This design decouples logical agents from physical model parameters, enabling full parameter sharing, full separation, and partial sharing within one interface, while allowing rewards to be assigned at role, turn, and trajectory levels.

At the system level, UnityMAS-O extends \texttt{verl} with a Ray-based star-topology runtime. A central controller executes the user-defined workflow, invokes tools or environments, records structured trajectories, and assembles rewards, while model-local worker groups perform rollout, buffering, advantage computation, and distributed PPO-style policy updates. The framework is task-agnostic: users can define the workflow, agents, model mapping, and rewards for different multi-agent systems without rewriting the underlying optimization infrastructure.

We instantiate UnityMAS-O on retrieval-augmented QA, iterative agentic search, and reflective code generation as representative workflow families. Across Natural Questions, HotpotQA, and held-out code tasks, multi-agent RL improves the same manually specified workflows before and after optimization, with especially large gains for smaller models and strict code all-passed metrics. These results demonstrate that UnityMAS-O can serve as a reusable optimization substrate for converting diverse LLM-based multi-agent workflows into trainable multi-agent RL systems.

\end{abstract}

\section{Introduction}

Large language models (LLMs) have made it possible to build multi-agent systems in which role-specialized agents communicate through natural language and tools to solve complex tasks~\citep{Vaswani+2017,brown2020language,li2023camel,wu2023autogen,qian2024chatdev}. A typical LLM-based multi-agent system defines a workflow over roles such as planners, retrievers, executors, critics, coders, and answerers. This decomposition is attractive because many real tasks are naturally procedural: information must be searched, filtered, verified, revised, and finally synthesized.

Despite this promise, the optimization interface of current multi-agent LLM systems remains limited. Most workflows are manually specified through prompts, routing rules, and hand-designed interaction protocols. When training is introduced, it is often applied to a single model or a single role, leaving the broader interaction pattern largely fixed. This creates a mismatch: the behavior that makes a multi-agent system succeed is not only the quality of an individual response, but also whether upstream agents produce useful intermediate states, whether downstream agents can use them, and whether rewards can be assigned to the roles that influence final outcomes.

Recent post-training frameworks have made reinforcement learning for LLMs much more practical by providing scalable rollout, optimization, and distributed execution infrastructure. TRL~\citep{vonwerra2020trl}, OpenRLHF~\citep{hu2024openrlhf}, slime~\citep{lmsys2025slime}, and verl/HybridFlow~\citep{sheng2024hybridflow} offer important building blocks for large-scale LLM training. However, these systems are primarily organized around single-policy optimization. Multi-turn interaction, tool use, or agentic behavior can be incorporated through customized environments, but the framework abstraction is still usually centered on one trainable policy rather than on a graph of interacting logical agents.

This limitation reflects a missing abstraction. A general framework for multi-agent LLM optimization should jointly represent four objects: the user-defined workflow graph, the logical agent roles in that graph, the mapping from logical agents to physical trainable models, and the reward functions defined over structured multi-agent trajectories. It should also provide the systems machinery needed to execute and optimize such workflows at scale. Without this abstraction, each multi-agent training pipeline becomes a task-specific implementation, making it difficult to compare parameter-sharing regimes, assign credit across roles, or reuse training infrastructure across workflows.

To address this gap, we present \textbf{UnityMAS-O}, a general-purpose optimization framework for LLM-based multi-agent systems developed by extending \texttt{verl}~\citep{sheng2024hybridflow}. The name reflects the goal of a \emph{unified} optimization interface for \emph{multi-agent systems}: ``Unity'' emphasizes a common abstraction across workflows, agents, model mappings, and rewards, while ``O'' denotes optimization. UnityMAS-O provides a general optimization interface for user-defined LLM multi-agent workflows. We describe its abstractions, system runtime, reward interface, and demonstrate its effectiveness on QA/search and code-generation workflows. A central controller executes the workflow, records graph-structured trajectories, invokes tools or environments, and assembles role-specific rewards. Model-local worker groups then perform inference, buffering, advantage computation, and distributed optimization for the physical LLM instances assigned to the logical agents.

The framework is built around three design requirements.

First, UnityMAS-O must optimize over multi-agent trajectories rather than isolated input--output pairs. The relevant learning signal may depend on intermediate evidence, verifier scores, tool outputs, or state updates produced by different agents at different stages. The framework therefore records the structured execution process and exposes rewards at the agent and workflow levels.

Second, it must decouple \emph{logical agents} from \emph{physical model instances}. The same workflow should support full parameter sharing, full parameter separation, and partial sharing without rewriting the training pipeline. This is essential for studying the trade-offs among specialization, coordination, stability, and computational cost.

Third, it must support heterogeneous distributed execution. Multi-agent workflows may invoke different LLMs, tools, and environments within one trajectory. UnityMAS-O therefore separates workflow-level control from model-local optimization, allowing different worker groups to manage different trainable models while the central controller preserves global trajectory and reward consistency.

We evaluate UnityMAS-O on two families of verifiable tasks: retrieval-augmented QA and reflective code generation. The QA experiments cover three search workflows on Natural Questions and HotpotQA, while the code experiments study a three-round plan--code--verify--reflect workflow with executable tests. Across these settings, MARL training improves the same workflow before and after optimization. The gains are especially pronounced for smaller QA models and strict code all-passed evaluation, and the trained code workflow also uses fewer verification turns on average. These results show that UnityMAS-O can instantiate different workflow families and improve them through multi-agent RL.

\paragraph{Contributions.}
This report makes the following contributions:
\begin{itemize}[leftmargin=1.5em]
    \item We present \textbf{UnityMAS-O}, a general-purpose optimization framework that extends \texttt{verl} from single-policy post-training to MARL over LLM-based multi-agent workflows.
    
    \item We introduce a unified abstraction that jointly represents user-defined workflow graphs, logical agent roles, agent-to-model mappings, role-specific rewards, and structured multi-agent trajectories.
    
    \item We provide native support for \textbf{full parameter sharing}, \textbf{full parameter separation}, and \textbf{partial parameter sharing}, enabling controlled study of specialization, resource cost, and optimization behavior.
    
    \item We design a Ray + \texttt{verl}-based distributed runtime with central workflow control, asynchronous rollout, reward assembly, model-local buffering, advantage computation, and per-model policy updates.
    
    \item We instantiate UnityMAS-O on QA/search and code workflows, showing before/after MARL gains, competitive parameter-shared training, and improved execution efficiency in reflective code generation.

    \item We release the open-source implementation at \url{https://github.com/chenyiqun/UnityMAS-O}.
\end{itemize}

Overall, UnityMAS-O is intended to serve as a practical systems substrate for training complex LLM-based multi-agent systems and as a platform for studying coordination, specialization, scalability, and optimization in such systems.

\section{Problem Statement}

We study how to optimize LLM-based multi-agent systems whose interaction structure is specified by the user. A workflow is represented as a directed computation graph
\[
\mathcal{G} = (\mathcal{V}, \mathcal{E}),
\]
where each node \(v \in \mathcal{V}\) denotes a \emph{logical agent role}, and each edge \(e \in \mathcal{E}\) specifies how information, control, or environment state can flow between roles. The graph may describe a sequential pipeline, a parallel retrieval stage, an iterative refinement loop, or a hybrid workflow that combines these patterns.

A key distinction is that \(\mathcal{G}\) defines the \emph{logical} multi-agent system, not necessarily the physical set of trainable models. A role may specify its prompt template, input/output interface, tool access, stopping condition, and reward-relevant behavior, but the workflow alone does not determine whether different roles share parameters, use separate LLMs, or use a partially shared configuration. The problem is therefore to optimize the behavior of the whole workflow while preserving this separation between logical roles and physical model instances.

\subsection{Inputs and Design Variables}

The framework takes as input a task distribution \(\mathcal{D}\), a workflow graph \(\mathcal{G}\), a set of trainable model instances
\[
\mathcal{M} = \{m_1, m_2, \dots, m_K\},
\]
and an agent--model mapping
\[
\phi: \mathcal{V} \rightarrow \mathcal{M}.
\]
The mapping \(\phi\) is a design variable rather than an implementation detail. It determines which physical LLM serves each logical role, and therefore controls the parameter-sharing regime of the multi-agent system:
\begin{itemize}[leftmargin=1.5em]
    \item \textbf{full sharing}, where all logical agents are mapped to a single model;
    \item \textbf{full separation}, where each logical agent is mapped to an independent model;
    \item \textbf{partial sharing}, where groups of related roles share parameters while other role groups use different models.
\end{itemize}
Different choices of \(\phi\) induce different trade-offs among specialization, coordination, parameter efficiency, memory cost, and training stability. A general optimization framework should make these choices explicit and configurable.

\subsection{Multi-Agent Trajectories and Optimization Objective}

For a task instance \(x \sim \mathcal{D}\), executing \(\mathcal{G}\) with mapping \(\phi\) induces a structured multi-agent trajectory
\[
\tau = (\mathcal{S}, \mathcal{A}, \mathcal{R}, \mathcal{C}),
\]
where \(\mathcal{S}\) contains workflow and environment states, \(\mathcal{A}\) contains outputs produced by the invoked agents, \(\mathcal{R}\) contains reward signals, and \(\mathcal{C}\) records the control-flow dependencies induced by the graph. Unlike a standard single-turn LLM trajectory, \(\tau\) may include retrieved evidence, tool outputs, partial solutions, verifier scores, memory updates, and repeated invocations of the same or different roles.

Rewards may be local to an agent invocation, delayed until the end of a sub-workflow, or defined only after the complete workflow terminates. The optimization objective is therefore not simply to maximize the quality of one response, but to maximize the expected return of the full multi-agent execution:
\[
\max_{\Theta} \;
\mathbb{E}_{x \sim \mathcal{D},\,
\tau \sim p_{\Theta}(\cdot \mid x, \mathcal{G}, \phi)}
\bigl[
R(\tau)
\bigr],
\]
where \(\Theta=\{\theta_m\}_{m\in\mathcal{M}}\) denotes the parameters of the trainable model instances, and \(R(\tau)\) may aggregate role-specific, turn-level, and trajectory-level rewards. This objective makes credit assignment central: the framework must connect final outcomes back to the agent actions and model parameters that produced the relevant intermediate states.

\subsection{Framework Requirements}

The framework problem is to provide a reusable training substrate for the objective above. Concretely, the framework should support:
\begin{itemize}[leftmargin=1.5em]
    \item graph-structured workflow specification with sequential, parallel, branching, and iterative execution;
    \item explicit agent--model mappings with full sharing, partial sharing, and full separation;
    \item agent-specific and global reward definitions, including delayed reward attribution over trajectories;
    \item asynchronous rollout, structured trajectory collection, advantage computation, and synchronized updates;
    \item heterogeneous distributed optimization over one or more LLM instances with potentially different sizes, optimizers, rollout backends, and hardware placements.
\end{itemize}
UnityMAS-O is designed to satisfy these requirements by exposing the agent-level abstractions in Section~\ref{sec:agent-framework} and the distributed training architecture in the subsequent system section.

\section{How UnityMAS-O Optimizes a Workflow}
\label{sec:how-it-works}

UnityMAS-O is designed around a simple user-facing workflow: users specify the multi-agent system they want to optimize, and the framework turns that specification into model-local RL training data. Concretely, optimization proceeds through the following steps.

\begin{enumerate}[leftmargin=1.5em]
    \item \textbf{Define logical agents.} The user specifies the roles in the workflow, such as planner, retriever, evidence extractor, coder, verifier, reflector, or answerer. Each role may have its own prompt template, output schema, tool access, and stopping condition.
    
    \item \textbf{Define the workflow graph.} The user connects roles into a graph that describes the execution process. The graph may be a sequential pipeline, a parallel retrieval stage, an iterative refinement loop, or a hybrid workflow.
    
    \item \textbf{Map agents to physical LLMs.} The user chooses an agent--model mapping. Logical agents may share one model, use independent models, or use a partially shared configuration.
    
    \item \textbf{Execute trajectories and assign rewards.} The central controller runs the workflow on task instances, records the graph-structured trajectory, invokes tools or environments, and evaluates role-, turn-, or trajectory-level rewards.
    
    \item \textbf{Update the mapped models.} Each model-local worker group receives the rollout fragments and rewards for the roles it served, constructs RL batches, computes advantages, and updates its assigned LLM.
\end{enumerate}

This interface separates the semantic design of the multi-agent workflow from the physical training layout. The same workflow can therefore be trained under different sharing regimes, reward designs, and distributed model placements.

\begin{table}[h]
\centering
\small
\setlength{\tabcolsep}{5pt}
\renewcommand{\arraystretch}{1.15}
\begin{tabularx}{\linewidth}{p{3.1cm}X X}
\toprule
Aspect & Standard \texttt{verl}-style RL & UnityMAS-O \\
\midrule
Optimization unit & One policy-centered prompt or environment trajectory & A graph-structured multi-agent workflow trajectory \\
Agent roles & Usually implicit in custom environment logic & First-class logical agents with role identities and interfaces \\
Model assignment & Centered on one trainable policy or manually customized model setup & Explicit agent--model mapping with full, partial, or independent sharing \\
Reward interface & Response-, token-, or trajectory-level rewards for one policy & Role-, turn-, and workflow-level rewards assigned to the responsible agents \\
Runtime structure & Rollout and update pipeline around model training & Central workflow controller plus model-local worker groups \\
\bottomrule
\end{tabularx}
\caption{Conceptual comparison between a standard \texttt{verl}-style RL setup and UnityMAS-O. UnityMAS-O keeps the scalable model-training machinery while adding workflow-level control, agent identities, reward attribution, and configurable agent--model mapping.}
\label{tab:verl-vs-unitymas}
\end{table}

\section{Agent Framework}
\label{sec:agent-framework}

\begin{figure}[t]
    \centering
    \includegraphics[width=\textwidth]{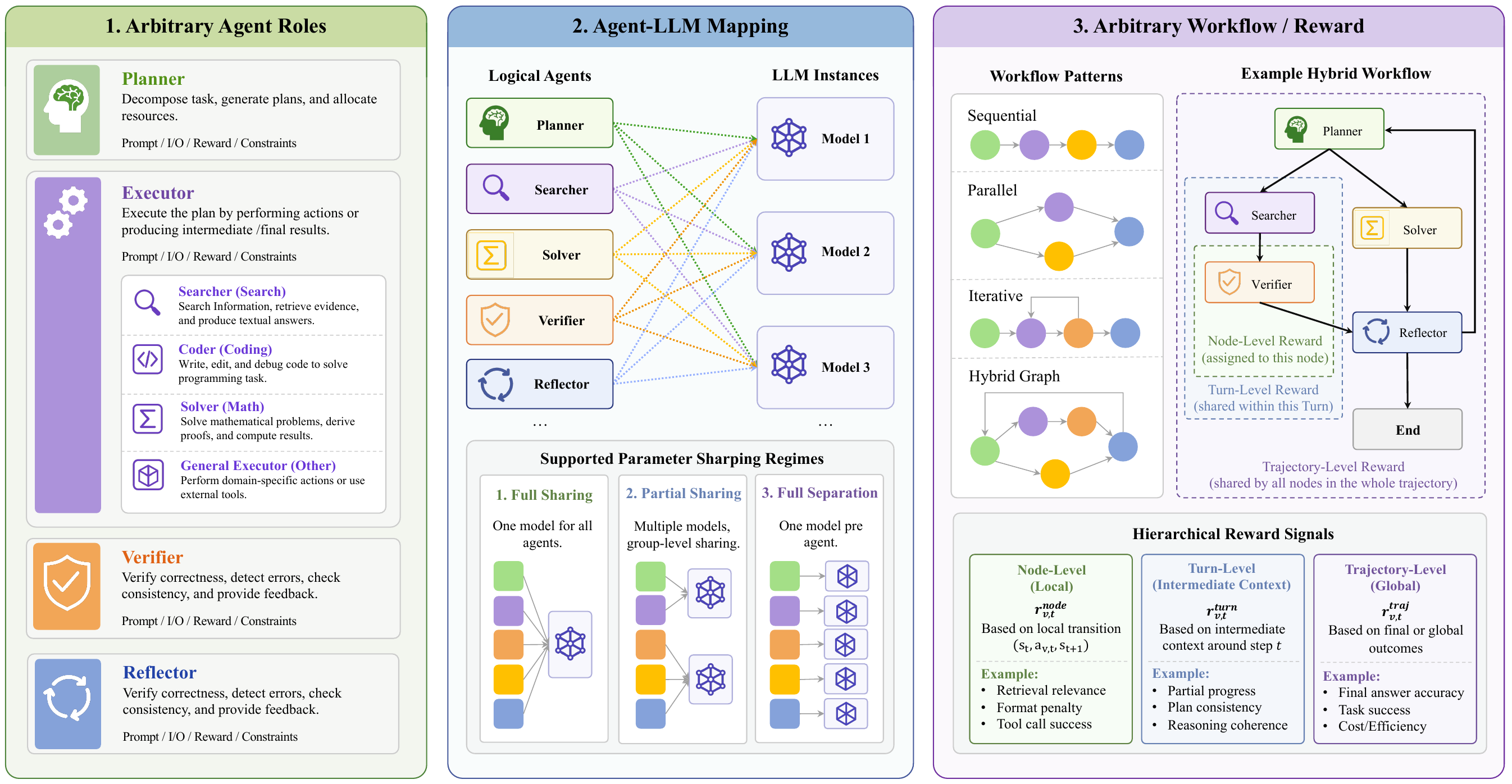}
    \caption{Agent-level abstractions in UnityMAS-O. Users define logical roles, map them to physical LLM instances, connect them with a workflow graph, and attach role-, turn-, or trajectory-level rewards for multi-agent RL optimization.}
    \label{fig:agent_framework}
\end{figure}

UnityMAS-O realizes the problem formulation above through four first-class abstractions: logical agent roles, agent--LLM mappings, workflow graphs, and reward functions. Figure~\ref{fig:agent_framework} gives the high-level view. These abstractions separate the semantic design of a multi-agent system from its physical parameterization and training runtime, allowing the same workflow to be executed under different sharing regimes, reward designs, and distributed model placements.

\subsection{Logical Agent Roles}

A logical agent role specifies what a node is responsible for inside a workflow. It may correspond to a planner, retriever, extractor, coder, verifier, reflector, summarizer, or answerer. Importantly, a role is a workflow-level object: it defines behavior and interfaces, not a unique parameter set.

We denote the set of roles as
\begin{equation}
\mathcal{V} = \{v_1, v_2, \dots, v_N\}.
\end{equation}
Each role may include its own prompt template, input/output schema, tool access, execution constraints, and reward-relevant behavior:
\begin{equation}
v_i = \bigl(\mathcal{P}_{v_i}, \mathcal{X}_{v_i}, \mathcal{Y}_{v_i}, \mathcal{T}_{v_i}, \mathcal{C}_{v_i}\bigr),
\end{equation}
where \(\mathcal{P}_{v_i}\) is the prompt template, \(\mathcal{X}_{v_i}\) and \(\mathcal{Y}_{v_i}\) are the input and output spaces, \(\mathcal{T}_{v_i}\) denotes available tools or environment interfaces, and \(\mathcal{C}_{v_i}\) captures execution constraints such as stopping conditions or required output formats.

This role abstraction is useful because the same logical role can be implemented by different LLMs, and multiple roles can also be implemented by the same LLM. The framework therefore optimizes role behavior without forcing a one-to-one correspondence between roles and parameters.

\subsection{Agent--LLM Mapping}

UnityMAS-O makes the mapping from logical roles to physical LLM instances explicit. Let
\begin{equation}
\phi: \mathcal{V} \rightarrow \mathcal{M}
\end{equation}
denote this mapping, where
\begin{equation}
\mathcal{M} = \{m_1, m_2, \dots, m_K\}
\end{equation}
is the set of trainable LLM instances.

The mapping \(\phi\) supports the main parameterization regimes used in multi-agent LLM systems:

\begin{itemize}[leftmargin=1.5em]
    \item \textbf{Full parameter sharing:} all logical agents are mapped to the same LLM instance:
    \[
    \forall v \in \mathcal{V},\quad \phi(v) = m_1.
    \]
    In this case, all agents share one policy model while being distinguished by their role prompts, inputs, outputs, and workflow positions.

    \item \textbf{Partial parameter sharing:} subsets of logical agents share parameters within each subset, while different subsets are mapped to different LLM instances. This allows related roles to share capacity while preserving specialization across role groups.

    \item \textbf{Full parameter separation:} each logical agent is assigned an independent LLM instance:
    \[
    \phi(v_i) = m_i.
    \]
    This setting maximizes role specialization but introduces higher memory, compute, and optimization costs.
\end{itemize}

By making \(\phi\) explicit, UnityMAS-O turns agent--model assignment into a configurable research object. This is the mechanism that enables controlled comparisons between shared and independent agents, and it also supports heterogeneous allocation in which different role groups use different model sizes, optimizers, rollout backends, or hardware placements.

\subsection{Workflow Graphs and Execution}

Given \(\mathcal{V}\), UnityMAS-O represents the workflow as
\begin{equation}
\mathcal{G} = (\mathcal{V}, \mathcal{E}),
\end{equation}
where edges in \(\mathcal{E}\) define the permissible flow of information, control, or environment state. The workflow graph is fully user-defined and can express sequential pipelines, parallel branches, iterative loops, and hybrid graph structures.

Execution on a task instance produces a structured trajectory. For notation, we can linearize the executed nodes as
\begin{equation}
\tau = \bigl(s_0, k_1, a_1, s_1, k_2, a_2, \dots, k_T, a_T, s_T\bigr),
\end{equation}
where \(s_t\) is the workflow state, \(k_t \in \mathcal{V}\) is the active role, and \(a_t\) is the output generated by that role. The state may include the user query, intermediate reasoning traces, retrieved evidence, tool outputs, partial answers, verifier feedback, memory states, and control-flow metadata. The linearized notation is only a view of the execution record; the dependency structure remains governed by \(\mathcal{G}\).

Conditioned on the current state and active role, the action is sampled from the LLM assigned by \(\phi\):
\begin{equation}
a_t \sim \pi_{\theta_{\phi(k_t)}}(\cdot \mid s_t, k_t),
\end{equation}
where \(\theta_{\phi(k_t)}\) denotes the parameters of the mapped LLM instance.

The next state is determined by workflow transition logic, tool execution, and environment feedback:
\begin{equation}
s_{t+1} \sim P(\cdot \mid s_t, a_t, k_t),
\end{equation}
where \(P\) is the workflow- and environment-dependent transition operator.

This formulation highlights the difference from standard single-turn post-training: UnityMAS-O optimizes the complete execution process of a graph-based multi-agent system.

\subsection{Agent-Specific Reward Interface}

UnityMAS-O does not impose one global reward function for all roles. Instead, each logical agent may define its own reward function according to its responsibility, position in the workflow, and available supervision signal.

For each role \(v \in \mathcal{V}\), we define
\begin{equation}
\mathcal{R}_v:
\mathcal{S} \times \mathcal{Y}_v \times \mathcal{S} \times \mathcal{Z}
\rightarrow \mathbb{R},
\end{equation}
where \(\mathcal{S}\) is the workflow state space, \(\mathcal{Y}_v\) is the output space of role \(v\), and \(\mathcal{Z}\) is the trajectory space. When role \(v\) is invoked, its reward is
\begin{equation}
r_v = \mathcal{R}_v(s, y_v, s', \tau).
\end{equation}

The reward can use node-level, turn-level, and trajectory-level information:
\begin{equation}
r_v
=
\mathcal{R}_v
\bigl(
\underbrace{s, y_v, s'}_{\text{node-level}},
\underbrace{h_t}_{\text{turn-level}},
\underbrace{\tau}_{\text{trajectory-level}}
\bigr),
\end{equation}
where node-level information describes the local input/output behavior of the current invocation, turn-level information captures the surrounding workflow context, and trajectory-level information reflects final or global workflow quality.

This interface covers rule-based format rewards, environment rewards, model-based rewards, and task metrics such as answer F1 or executable test pass rates. It also supports combinations of local and global supervision:
\begin{equation}
r_v
=
\lambda_v^{\mathrm{node}} r_v^{\mathrm{node}}
+
\lambda_v^{\mathrm{turn}} r_v^{\mathrm{turn}}
+
\lambda_v^{\mathrm{traj}} r_v^{\mathrm{traj}},
\end{equation}
where the coefficients are optional user-defined weights. This additive form is not required; users may define sparse, dense, delayed, non-additive, or task-specific attribution rules. The concrete QA and code reward definitions used in our experiments are described later in Section~\ref{sec:workflows-rewards}.

\subsection{Optimization Objective and System Interface}

Given \(\mathcal{V}\), \(\mathcal{G}\), \(\phi\), and the reward functions above, UnityMAS-O optimizes one or more trainable LLM instances over graph-structured trajectories. Let
\begin{equation}
\Theta = \{\theta_m\}_{m \in \mathcal{M}}
\end{equation}
be the collection of trainable parameters.

For a trajectory \(\tau\), let \(\mathcal{I}_v(\tau)\) denote the set of steps in which role \(v\) is invoked. The training objective can be written as
\begin{equation}
\max_{\Theta} \;
\mathbb{E}_{x \sim \mathcal{D},\,
\tau \sim p_{\Theta}(\cdot \mid x, \mathcal{G}, \phi)}
\left[
\sum_{v \in \mathcal{V}}
\sum_{t \in \mathcal{I}_v(\tau)}
\gamma^{t-1}
r_{v,t}
\right],
\end{equation}
where
\begin{equation}
r_{v,t}
=
\mathcal{R}_v(s_t, y_{v,t}, s_{t+1}, \tau)
\end{equation}
is the reward assigned to role \(v\) at step \(t\). The objective does not require all roles to optimize the same immediate signal; it only requires that rewards can be committed to the model instance that generated the corresponding action through \(\phi\).

The agent framework therefore defines the contract between user-specified multi-agent workflows and the training system: roles produce actions, the workflow records structured trajectories, rewards are attributed to role invocations, and \(\phi\) routes each training signal to the correct physical model. The next section describes how this contract is implemented at scale through asynchronous workflow execution, Ray-based scheduling~\citep{moritz2018ray}, model-local buffers, and PPO-style updates~\citep{schulman2017ppo}.

\section{System and Training}

\begin{figure}[t]
    \centering
    \includegraphics[width=\textwidth]{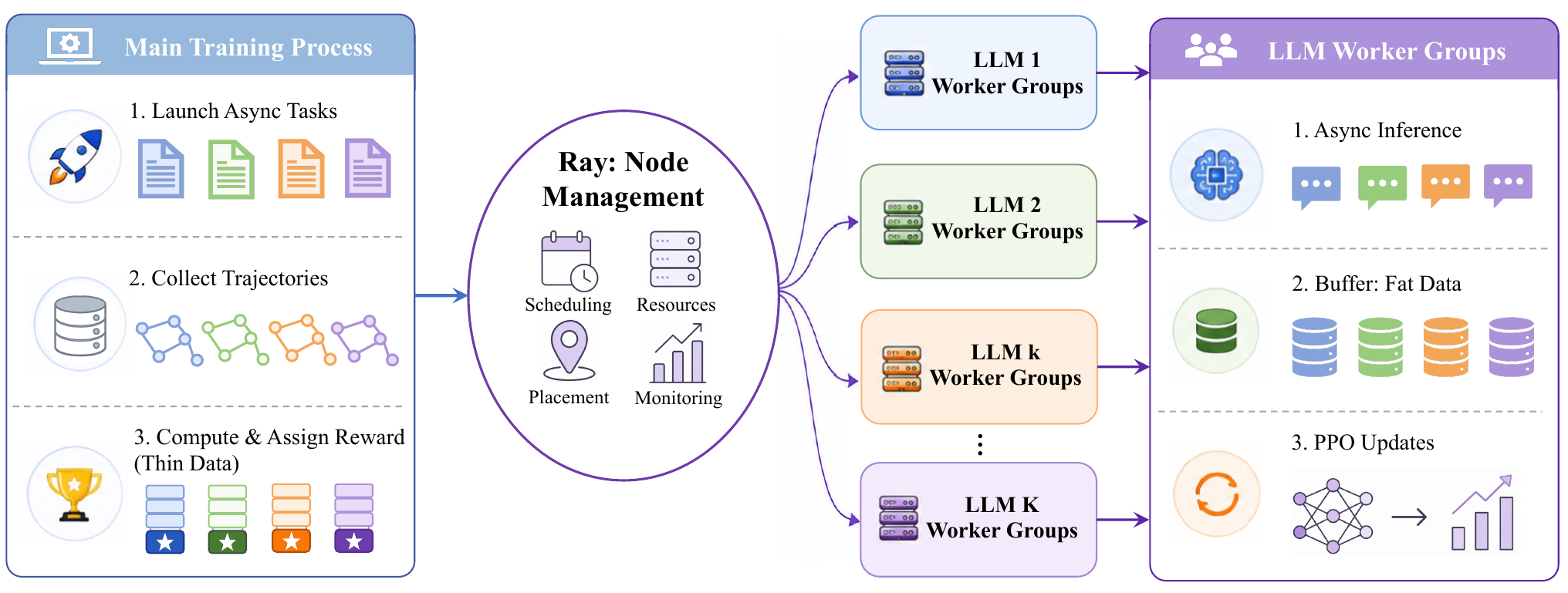}
    \caption{Distributed training architecture of UnityMAS-O. A central controller executes user-defined workflows, routes logical agent invocations to model-local worker groups, assembles rewards and trajectory metadata, and coordinates distributed policy updates.}
    \label{fig:opt_framework}
\end{figure}

The previous sections define the optimization problem and the agent-level abstractions. This section describes how UnityMAS-O makes them executable at scale. The central systems question is how to run a graph-structured multi-agent workflow while updating one or more physical LLMs according to the agent--model mapping \(\phi\). UnityMAS-O answers this with a star-topology runtime: a central controller owns workflow state and reward assignment, while model-local worker groups own generation, tensor buffering, advantage computation, and parameter updates.

\subsection{Runtime Architecture}

As shown in Figure~\ref{fig:opt_framework}, UnityMAS-O has three system components. The \textbf{central controller} maintains the global training loop, schedules active workflow states, evaluates stopping conditions, invokes tools or environments, assembles rewards, and coordinates updates. The \textbf{Ray execution layer} provides remote invocation, worker lifecycle management, and GPU placement across machines~\citep{moritz2018ray}. The \textbf{LLM worker groups} are model-local training units; each worker group is associated with one physical LLM instance and handles the agent invocations routed to that model.

Let \(\mathcal{M}=\{m_1,\dots,m_K\}\) be the trainable model instances. UnityMAS-O instantiates corresponding worker groups
\begin{equation}
\mathcal{W} = \{W_1, W_2, \dots, W_K\},
\end{equation}
where \(W_i\) serves model \(m_i\). The mapping \(\phi:\mathcal{V}\rightarrow\mathcal{M}\) determines routing. If role \(v\) is mapped to \(m_i\), then its generation requests, rollout tensors, reward assignments, and policy updates are handled by \(W_i\). This directly implements full sharing, partial sharing, and full separation: changing \(\phi\) changes the physical training layout without changing the logical workflow graph.

\subsection{Workflow Scheduling and Routing}

For a batch of task instances \(\{x^{(n)}\}_{n=1}^{N}\), the controller executes their workflow graphs asynchronously. At each scheduler step, it identifies the next executable role for each active trajectory. If the current workflow state is \(s_t\) and the active role is \(k_t\in\mathcal{V}\), the controller routes the generation request to the worker group selected by \(\phi\):
\begin{equation}
(s_t, k_t) \longrightarrow W_{\phi(k_t)}.
\end{equation}
The worker group returns a lightweight action output, such as generated text or a structured field. The controller then advances the workflow state, applies environment or tool transitions when needed, evaluates stopping conditions, and schedules the next executable role. This design exposes concurrency at two levels: multiple task instances can run in parallel, and multiple branches within a workflow can be dispatched concurrently when the graph permits.

\subsection{Model-Local Buffers}

A key systems principle in UnityMAS-O is to separate workflow-level control data from model-side training data. The controller only needs lightweight information: role identities, routing identifiers, generated outputs, workflow states, and reward metadata. Heavy tensors produced during generation, including token log probabilities, attention masks, value estimates, and rollout metadata, remain local to the worker group that produced them.

For an invocation of role \(v\) at step \(t\), where \(\phi(v)=m_i\), the resulting training fragment is committed to the buffer of \(W_i\):
\begin{equation}
(s_t, v, y_{v,t}, s_{t+1}, r_{v,t}) \longrightarrow \mathcal{B}_i,
\end{equation}
where \(\mathcal{B}_i\) is the model-specific buffer for \(m_i\). In practice, this logical transition is stored together with the token-level tensors needed for optimization. Keeping these tensors model-local reduces communication overhead and preserves the link between a generated action, its role identity, its reward, and the physical model that must be updated.

Once sufficient rollout data has accumulated, each buffer is converted into a ready-to-update batch:
\begin{equation}
\mathcal{B}_i \rightarrow \mathcal{D}_i^{\mathrm{ready}}.
\end{equation}
Ready-batch construction may include filtering incomplete trajectories, aligning delayed rewards with token sequences, padding, shuffling, and matching the data-parallel configuration of the corresponding worker group. Because batches are built per model instance, different worker groups can receive different amounts of data and still update under one global workflow.

\subsection{PPO-Style Training Loop}

UnityMAS-O is designed to be compatible with different RL algorithms, while the current implementation follows a PPO-style training loop~\citep{schulman2017ppo}. A training iteration proceeds as follows.

\paragraph{Asynchronous rollout.}
The controller launches workflow executions and dispatches each role invocation to the mapped worker group. Worker groups generate outputs, return lightweight actions to the controller, and keep model-side rollout tensors in local buffers.

\paragraph{Reward assembly and commitment.}
Rewards are evaluated after an agent invocation, a sub-workflow, or a full trajectory depending on the user-defined reward function. The controller attaches each reward to the corresponding role invocation and commits it to the buffer of the model that served that role.

\paragraph{Batch construction.}
The controller and worker groups construct model-specific ready batches. This step aligns trajectory-level or delayed rewards with the appropriate token sequences and preserves role identifiers for logging, attribution, and role-specific analysis.

\paragraph{Advantage computation and policy evaluation.}
Each worker group computes the quantities required for PPO-style optimization, including old log probabilities, reference-policy log probabilities, value estimates, returns, and advantages. For example, generalized advantage estimation~\citep{schulman2015gae} computes
\begin{equation}
A_t =
\sum_{l=0}^{T-t}
(\gamma \lambda)^l \delta_{t+l},
\end{equation}
where
\begin{equation}
\delta_t = r_t + \gamma V(s_{t+1}) - V(s_t).
\end{equation}

\paragraph{Model-specific updates and synchronization.}
Each worker group updates its own actor and critic parameters using its ready batch. Different groups may use different learning rates, optimizer settings, precision modes, gradient accumulation schedules, or parallelization strategies. After the actor update, updated weights are synchronized to the rollout backend, such as vLLM~\citep{kwon2023vllm}, so subsequent workflow executions use the latest policy.

\subsection{Parameter Sharing in the System}

The same runtime supports all parameter-sharing regimes introduced in Section~\ref{sec:agent-framework}. In the fully shared setting, all roles route to one worker group, and their trajectory fragments jointly update one model. Role identities are still preserved, so UnityMAS-O can compute role-specific rewards and report agent-level metrics even when the physical parameters are shared.

In the fully separated setting, different roles route to different worker groups. Each model receives only the data generated by the roles it serves, which encourages specialization but requires more memory and computation. In the partially shared setting, related roles share a worker group while other roles use separate models. Thus, UnityMAS-O can change the resource--specialization trade-off by changing \(\phi\), without rewriting workflow logic or reward definitions.

\subsection{Summary}

The system layer completes the bridge from the formal objective to an executable training framework. The controller maintains graph-structured workflow execution and reward attribution; Ray manages distributed remote execution and resource placement; worker groups perform model-local generation, buffering, advantage computation, PPO-style updates, and weight synchronization. This division of responsibility is what allows UnityMAS-O to optimize user-defined multi-agent workflows under shared, partially shared, or fully independent LLM parameterizations.

\begin{figure}[t]
    \centering
    \includegraphics[width=\textwidth]{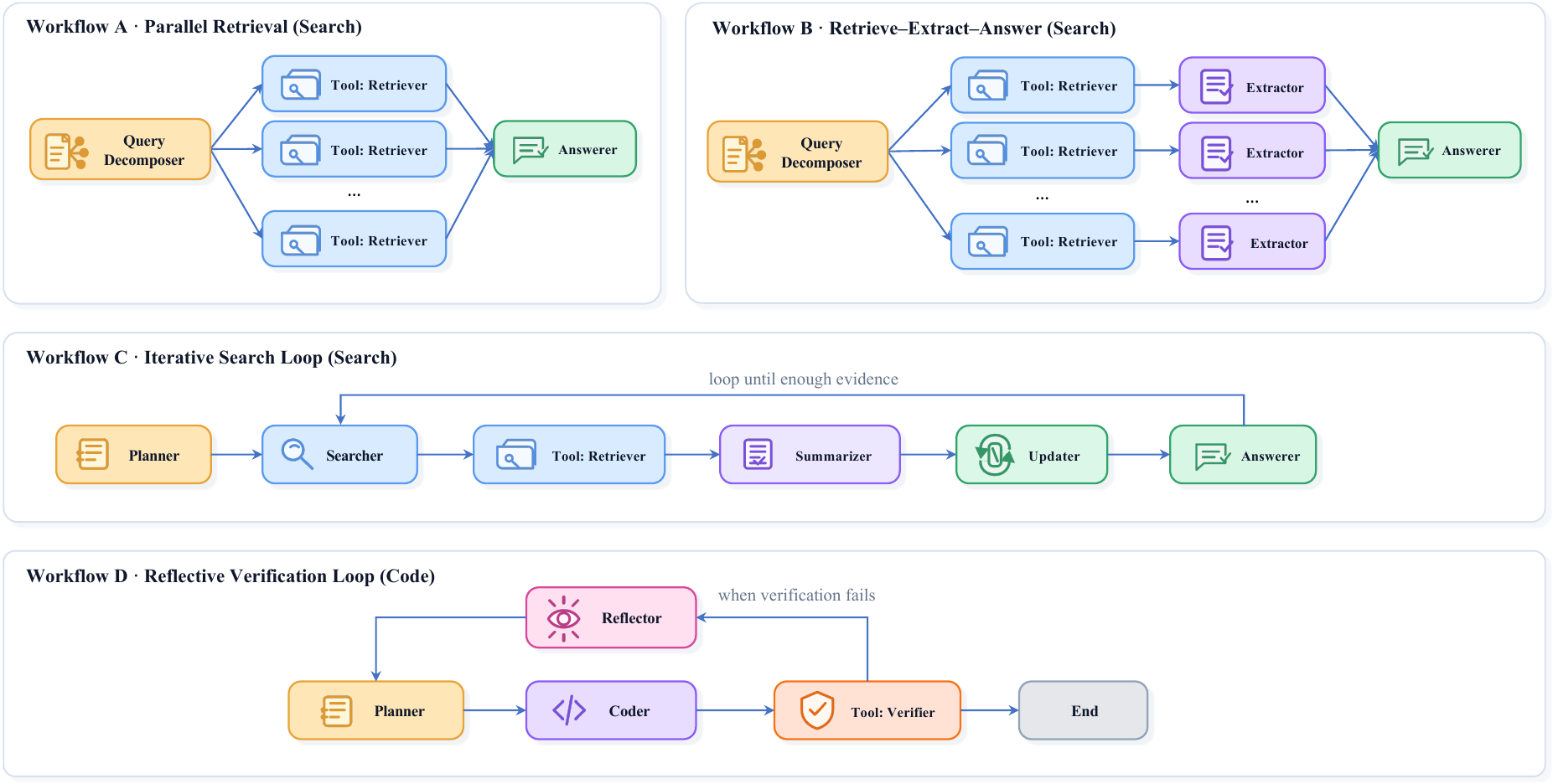}
    \caption{Workflow templates used in the experiments. The figure shows three search workflows---parallel retrieval, retrieve--extract--answer, and iterative search---and one code workflow with reflective verification.}
    \label{fig:opt_workflow}
\end{figure}

\section{Workflows and Rewards}
\label{sec:workflows-rewards}

This section summarizes the workflow instances used in our experiments and the corresponding reward assignment rules. Figure~\ref{fig:opt_workflow} illustrates the four workflow templates, and Table~\ref{tab:workflow-settings} summarizes their agent composition, model mapping, and reward signals. The goal is to make the optimization target explicit: UnityMAS-O does not only execute a multi-agent graph, but also assigns rewards to the individual trainable agent nodes according to their role in the workflow.

\begin{table}[t]
\centering
\small
\caption{Workflow settings evaluated in UnityMAS-O. ``LLM agents'' counts trainable logical agents; tools such as retrievers, context builders, code execution, and answer graders are not counted as trainable agents.}
\label{tab:workflow-settings}
\resizebox{\linewidth}{!}{
\begin{tabular}{p{2.7cm}p{4.1cm}p{1.6cm}p{4.2cm}p{3.9cm}}
\toprule
Domain & Workflow & LLM agents & Model mapping & Reward / validation signal \\
\midrule
Search & Workflow A: Parallel Retrieval & 2 & decompose and answer use separate model groups & shared final-answer F1 plus node-level format penalties \\
Search & Workflow B: Retrieve--Extract--Answer & 3 & decompose, evidence, and answer use separate model groups; evidence calls share one evidence model & shared final-answer F1 plus node-level format penalties \\
Search & Workflow C: Iterative Search Loop & 5 & planning and answer share a reasoning model; search, summary, and update use separate models & M-ASK turn-level F1 rewards \\
Code & Workflow D: Reflective Verification Loop & 3 & planner, coder, and reflector use separate model groups; code verifier is a tool & verifier-score and score-delta rewards over three rounds \\
\bottomrule
\end{tabular}}
\end{table}

\paragraph{Workflow A: Parallel Retrieval (Search).}
This workflow decomposes the input question into retrieval queries, retrieves evidence passages in parallel, and then generates the final answer. Let \(V_{\mathrm{A}}\) denote the set of trainable agent nodes in the workflow, and let \(r^{\mathrm{fmt}}_v\) be the format penalty for node \(v\). After the answer node produces \(\hat{y}\), we compute the normalized answer score \(F_1(\hat{y}, y)\) against the ground truth \(y\). The same final F1 score is shared by every trainable agent node:
\[
r_v = F_1(\hat{y}, y) + r^{\mathrm{fmt}}_v, \quad v \in V_{\mathrm{A}}.
\]
Thus, each agent is locally constrained by its own output format while being globally optimized for final answer quality.

\paragraph{Workflow B: Retrieve--Extract--Answer (Search).}
This workflow extends Workflow A by inserting evidence extraction agents between retrieval and answer generation. The reward rule is the same as Workflow A. Each trainable agent node, including query decomposition, evidence extraction, and final answer generation, receives its own format penalty. Once the final answer is produced, the final answer F1 score is shared by all trainable agent nodes:
\[
r_v = F_1(\hat{y}, y) + r^{\mathrm{fmt}}_v, \quad v \in V_{\mathrm{B}}.
\]
The shared F1 reward encourages the evidence extraction nodes to preserve information useful for the final answer, while the node-level format penalty prevents invalid intermediate structures.

\paragraph{Workflow C: Iterative Search Loop (Search).}
For the iterative search workflow, we follow the M-ASK turn-level reward design~\citep{chen2026mask}. The planning agent first constructs an initial knowledge state and an initial answer \(a_0\). Its reward is the absolute answer quality:
\[
r_{\mathrm{plan}} = F_1(a_0, y).
\]
During training, the answer agent is invoked after each search/update turn as an intermediate evaluator. At turn \(t\), it generates a temporary answer \(a_t\), and the answer agent receives the absolute reward
\[
r_{\mathrm{ans}}^{(t)} = F_1(a_t, y).
\]
The collaborative loop agents, i.e., search, summary, and update, share the marginal improvement of answer quality:
\[
r_{\mathrm{iter}}^{(t)} = F_1(a_t, y) - F_1(a_{t-1}, y).
\]
This reward is assigned to the search, summary, and update agents active in turn \(t\). If the search agent outputs \texttt{<end>}, its task reward is set to 0. This makes the iterative loop optimize step-wise information gain rather than only the final sparse outcome.

\paragraph{Workflow D: Reflective Verification Loop (Code).}
The code workflow is unrolled for three verification rounds:
\[
\begin{aligned}
&\mathrm{Planner}_0 \rightarrow \mathrm{Coder}_0 \rightarrow \mathrm{Verifier}_0(s_0) \rightarrow \mathrm{Reflector}_0, \\
&\mathrm{Planner}_1 \rightarrow \mathrm{Coder}_1 \rightarrow \mathrm{Verifier}_1(s_1) \rightarrow \mathrm{Reflector}_1, \\
&\mathrm{Planner}_2 \rightarrow \mathrm{Coder}_2 \rightarrow \mathrm{Verifier}_2(s_2).
\end{aligned}
\]
Here \(s_i\) is the verifier score after round \(i\), such as the test pass rate. The verifier is a non-trainable tool node, while planner, coder, and reflector are trainable agents. The task rewards are assigned as:
\[
r(\mathrm{Planner}_0) = r(\mathrm{Coder}_0) = s_0,
\]
\[
r(\mathrm{Reflector}_0) = r(\mathrm{Planner}_1) = r(\mathrm{Coder}_1) = s_1 - s_0,
\]
\[
r(\mathrm{Reflector}_1) = r(\mathrm{Planner}_2) = r(\mathrm{Coder}_2) = s_2 - s_1.
\]
This design credits the first planner--coder pair for the initial solution quality, and then credits each reflector and subsequent planner--coder pair only for the improvement it creates over the previous verified solution.

\section{Experiments}
\label{sec:experiments}

\subsection{Experimental Setup}

We evaluate UnityMAS-O on two families of verifiable tasks that require different forms of multi-agent coordination: retrieval-augmented question answering and iterative code generation. For QA, we use Natural Questions (NQ)~\citep{kwiatkowski-etal-2019-natural} and HotpotQA~\citep{yang2018hotpotqa}; NQ mainly tests single-hop open-domain retrieval, while HotpotQA stresses multi-hop evidence aggregation. For code generation, we follow the DeepCoder data setting, which uses approximately 24K unique programming problem--test pairs compiled from TACO-Verified, PrimeIntellect SYNTHETIC-1, and LiveCodeBench v5 problems from 2023-05-01 to 2024-07-31~\citep{li2023taco,jain2024livecodebench,deepcoder2025}. Each instance is verifiable through executable tests, making it suitable for the verifier-based reward used in Workflow D. Unless otherwise specified, the trainable agents use Qwen3-family backbones~\citep{yang2025qwen3}.

The QA metric is normalized answer F1. For code, we use held-out test all-passed rate as the main evaluation metric and report the training final pass rate as a learning-dynamics signal. For every task, workflow, and model-size/mapping setting, we compare performance before multi-agent RL training with the best performance achieved during training. The goal of this report is not to exhaustively benchmark all possible agentic systems, but to show that UnityMAS-O can instantiate different workflow families and improve them through multi-agent RL.

\subsection{QA and Agentic Search Results}

\begin{figure}[t]
\centering
\includegraphics[width=\linewidth]{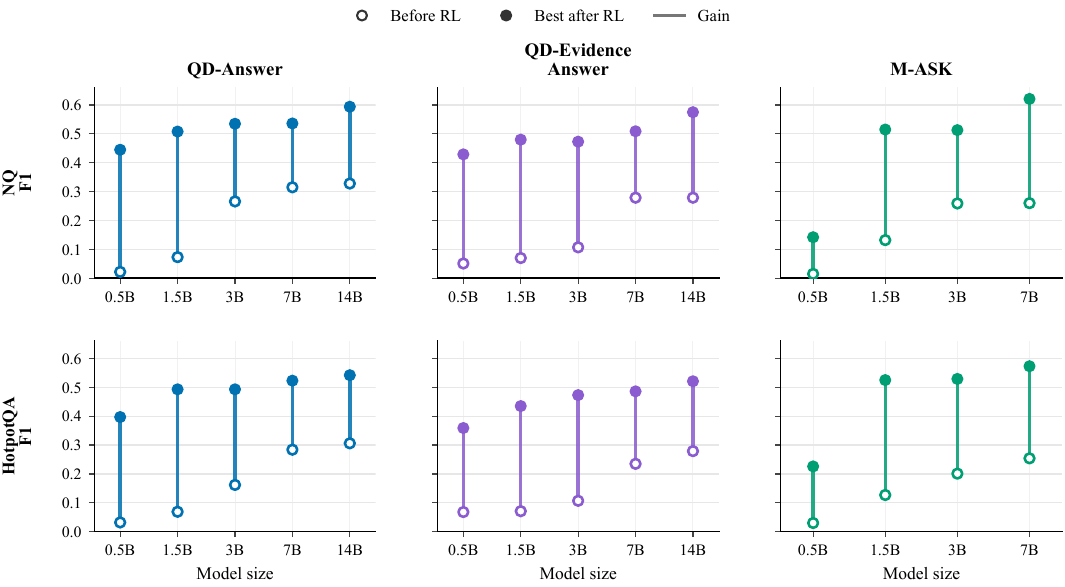}
\caption{Effect of multi-agent RL training on QA validation F1 across datasets, workflows, and model scales. Each vertical segment connects performance before RL training to the best validation performance achieved during training; open markers denote before-RL performance and filled markers denote best-after-RL performance.}
\label{fig:qa-training-gains}
\end{figure}

The QA experiments illustrate how UnityMAS-O optimizes agentic search systems whose agents coordinate through intermediate query, evidence, and answer states. Table~\ref{tab:qa-results} reports the best validation F1 achieved after training, while Figure~\ref{fig:qa-training-gains} shows the before/after gains induced by MARL. The key observation is that the gains are broad rather than isolated: every QA workflow and model scale we evaluate improves after training, even though the workflows expose different role structures and communication paths.

The gains are especially large for small models, where the untrained multi-agent workflows often begin close to failure. For example, QD-Retrieve-Answer improves from 0.022 to 0.445 on NQ with 0.5B agents, and from 0.032 to 0.397 on HotpotQA; full numeric values are provided in Table~\ref{tab:qa-gains-full}. The 0.5B and 1.5B evidence workflows also move from weak initial behavior to useful validation performance. These improvements indicate that MARL is not merely selecting a better checkpoint from an already competent policy. Instead, training helps role-specialized agents satisfy workflow protocols, preserve answer-bearing evidence, and coordinate toward final answer quality.

The workflow-level differences should be read as examples of what MARL can optimize, rather than as a search for a universally best workflow. On NQ, many questions can be answered through a relatively direct retrieval path, so QD-Retrieve-Answer remains strong after training and reaches 0.594 F1 at 14B. On HotpotQA, which requires multi-hop evidence aggregation, M-ASK benefits more from its iterative state updates and becomes the strongest workflow at 1.5B, 3B, and 7B. This pattern is consistent with the central mechanism of UnityMAS-O: when a workflow exposes meaningful intermediate decisions, MARL can assign turn-level credit to those decisions and improve the behavior of the multi-agent system as a whole.

\begin{table}[t]
\centering
\small
\caption{Peak validation F1 for QA workflows. Bold marks the best value under the same task and model-size column. Entries correspond to the model-size/mapping used by each workflow, e.g., 2x for two-agent RAG, 3x for evidence RAG, and 4x for M-ASK.}
\label{tab:qa-results}
\resizebox{\linewidth}{!}{
\begin{tabular}{llcccccc}
\toprule
Task & Workflow & 0.5B & 1.5B & 3B & 7B & 14B \\
\midrule
\multirow{3}{*}{NQ}
& QD-Retrieve-Answer & \textbf{0.445} & 0.508 & \textbf{0.535} & 0.536 & \textbf{0.594} \\
& QD-Retrieve-Evidence-Answer & 0.429 & 0.480 & 0.473 & 0.509 & 0.575 \\
& M-ASK iterative search & 0.142 & \textbf{0.515} & 0.513 & \textbf{0.621} & -- \\
\midrule
\multirow{3}{*}{HotpotQA}
& QD-Retrieve-Answer & \textbf{0.397} & 0.493 & 0.493 & 0.523 & \textbf{0.542} \\
& QD-Retrieve-Evidence-Answer & 0.359 & 0.435 & 0.473 & 0.486 & 0.522 \\
& M-ASK iterative search & 0.226 & \textbf{0.525} & \textbf{0.529} & \textbf{0.573} & -- \\
\bottomrule
\end{tabular}}
\end{table}

\begin{figure}[t]
    \centering
    \includegraphics[width=\linewidth]{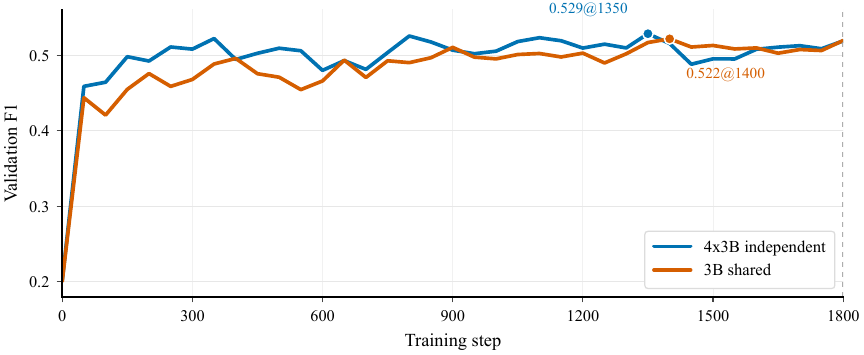}
    \caption{HotpotQA M-ASK validation F1 for a 3B shared-parameter setting and a 4x3B independent setting over a matched 0--1800 training interval.}
    \label{fig:mask-3b-sharing}
\end{figure}

\paragraph{Parameter sharing in M-ASK.}
Figure~\ref{fig:mask-3b-sharing} further illustrates the agent--model decoupling design in UnityMAS-O. It compares the HotpotQA M-ASK 3B shared run with the 4x3B independent run over the matched 0--1800 training interval. Both runs start from almost the same F1 (0.200 vs. 0.201). The independent setting climbs slightly faster and reaches 0.529 F1 by step 1350, while the shared setting reaches 0.522 F1 by step 1400 and 0.520 at step 1800. Thus, within the aligned training window, parameter sharing converges more slowly but reaches a similar level. This suggests that UnityMAS-O can train logical multi-agent workflows even when several roles share the same physical LLM parameters, reducing the number of model groups while preserving most of the final task performance.

\subsection{Code Results}

\begin{figure}[t]
\centering
\includegraphics[width=\linewidth]{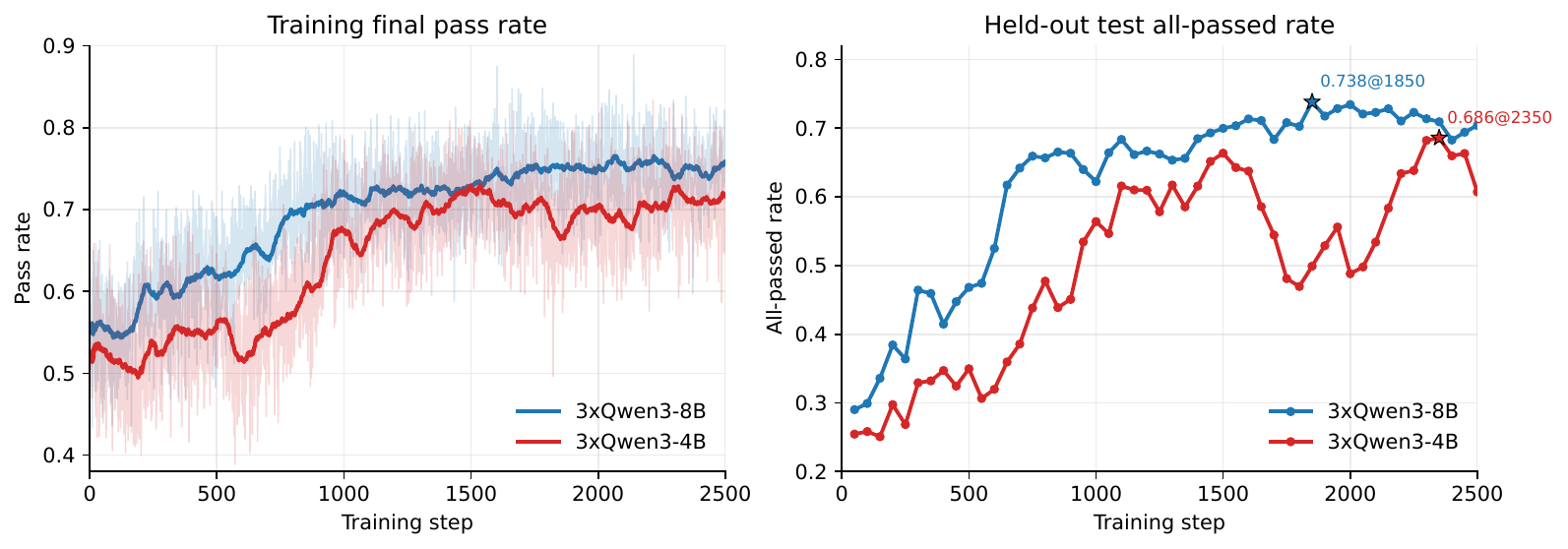}
\caption{Training dynamics of the Reflective Verification Loop code workflow up to step 2500. The left panel shows training final pass rate with a moving average for readability, and the right panel shows held-out test all-passed rate at validation checkpoints. Peak markers indicate the best held-out all-passed rates within the plotted range.}
\label{fig:code-train-test-curves}
\end{figure}

\begin{figure}[t]
\centering
\includegraphics[width=\linewidth]{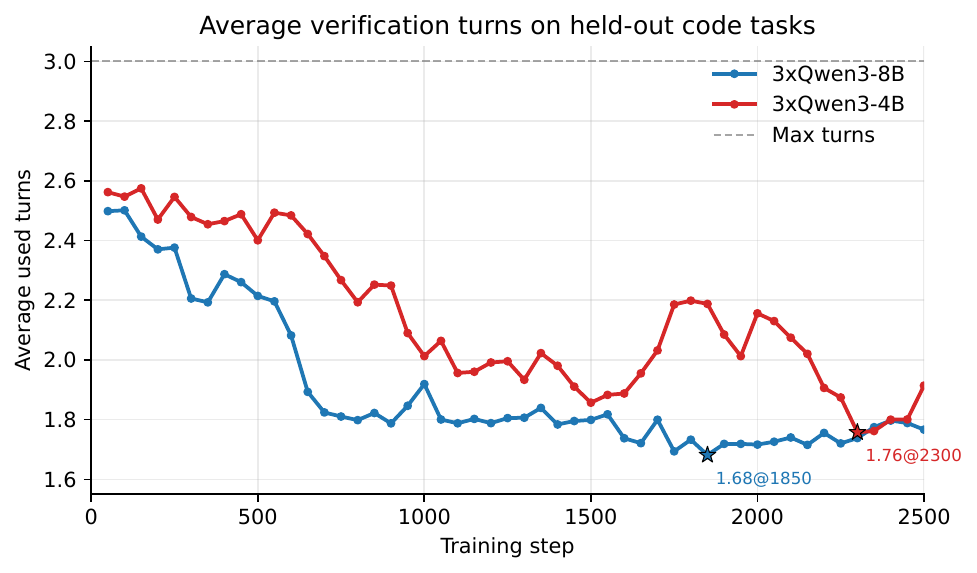}
\caption{Average number of verification turns used by the Reflective Verification Loop on held-out code examples. The workflow allows at most three verification rounds and terminates early once the generated solution passes all verifier tests.}
\label{fig:code-used-turns}
\end{figure}

The code workflow provides a complementary demonstration because success depends on a closed-loop plan--code--verify--reflect process. The verifier executes candidate programs and returns pass-rate feedback for training, while held-out evaluation measures whether the final program passes all tests. Table~\ref{tab:code-results} reports the compact before/after comparison under this all-passed metric. Figure~\ref{fig:code-train-test-curves} shows the training diagnostic and held-out evaluation trajectories, and Figure~\ref{fig:code-used-turns} measures how many verification rounds are used before early termination.

\begin{table}[t]
\centering
\caption{Effect of multi-agent RL training on the Reflective Verification Loop code workflow, measured by held-out test all-passed rate.}
\label{tab:code-results}
\footnotesize
\setlength{\tabcolsep}{10pt}
\renewcommand{\arraystretch}{1.08}
\begin{tabular}{lcccc}
\toprule
Setting & Before RL & Best after RL & Abs. gain & Rel. gain \\
\midrule
3xQwen3-4B & 0.255 & 0.686 & +0.431 & +169\% \\
3xQwen3-8B & 0.290 & \textbf{0.738} & +0.448 & +154\% \\
\bottomrule
\end{tabular}
\end{table}

For code generation, MARL training produces large gains under the stricter all-passed evaluation. The 3xQwen3-4B workflow improves from 0.255 to 0.686, a +0.431 absolute gain, while the 3xQwen3-8B workflow improves from 0.290 to 0.738, a +0.448 absolute gain. These improvements are substantial because the metric requires the final generated program to pass the full held-out test suite, rather than rewarding partially correct solutions. The result shows that UnityMAS-O can train a multi-agent code workflow end to end: the planner and coder are rewarded for stronger candidate solutions, while the reflector is rewarded when its feedback improves the next verified attempt.

Figure~\ref{fig:code-train-test-curves} separates the optimization diagnostic from the main held-out metric. The training pass-rate curves generally increase over training, but we do not interpret their exact maxima because minibatch-level pass rates are visibly noisy. The held-out all-passed curves provide the main evaluation signal within the plotted 0--2500 step range: the 3xQwen3-8B setting reaches its best held-out score of 0.738 at step 1850, whereas the 3xQwen3-4B setting continues improving over a longer horizon and peaks at 0.686 at step 2350. This trajectory is consistent with MARL improving the interaction pattern of the workflow, not only the final generation step. 

The used-turns curve in Figure~\ref{fig:code-used-turns} shows that the accuracy gains are accompanied by more efficient execution. Because the workflow stops once a solution passes all tests, lower used-turn counts mean that more examples are solved before exhausting the three-round budget. Both settings start near 2.5 turns on average. Within the 0--2500 step range, the 3xQwen3-8B workflow reaches a minimum of 1.68 turns at step 1850, while the 3xQwen3-4B workflow reaches 1.76 turns near step 2300 and uses 1.76 turns at its best all-passed checkpoint. These values are roughly 40\% below always using all three rounds, suggesting that MARL improves not only final correctness but also the internal stopping behavior of the reflective workflow.

\subsection{Cross-Workflow Analysis}

Across QA and code, the experiments show the same high-level pattern: once a multi-agent workflow is specified, UnityMAS-O can turn it into a trainable multi-agent RL problem and improve its behavior. The initial policies often fail to coordinate. QA agents may violate output formats, retrieve evidence without preserving answer-bearing content, or pass incomplete intermediate states to the next role; code agents may require multiple verifier--reflection rounds before producing a correct program. After MARL training, the same logical workflows become much stronger, as shown by the broad before/after gains in Figure~\ref{fig:qa-training-gains} and the strict all-passed improvements in Table~\ref{tab:code-results}.

The gains are useful because they occur inside multi-agent workflows rather than only at the final response. In QA, MARL improves not only the final answerer but also the upstream roles that determine what information reaches the answerer. In code, the planner, coder, and reflector are trained through verifier-grounded feedback, so improvement depends on temporal coordination across turns rather than isolated better code generation.

These results illustrate the core use case of UnityMAS-O: multi-agent LLM systems can be optimized at the level of roles, communication, and credit assignment. The reward design provides credit to the agents whose decisions affect downstream outcomes, including final-answer rewards for search workflows and verifier-score deltas for the reflective code workflow. The parameter-sharing result shows that this optimization can be applied even when logical agents do not map one-to-one to separate model groups, and the used-turns result shows that MARL can improve efficiency rather than simply spending more computation. Together, the QA and code results demonstrate that UnityMAS-O converts manually specified workflows into trainable multi-agent policies and improves correctness, resource flexibility, and execution efficiency in the evaluated settings.

Beyond the experiments reported here, we are also conducting ongoing MARL experiments with UnityMAS-O on additional workflow families, including embodied-agent tasks such as ALFWorld~\citep{shridhar2021alfworld}, web-interaction tasks such as WebShop~\citep{yao2022webshop}, and software-engineering tasks such as SWE-bench~\citep{jimenez2024swebench}. These experiments are not yet complete and are therefore not included in the quantitative results of this report. Their purpose is to further validate the same framework interface under more diverse task structures, tool interfaces, trajectory lengths, and reward definitions.

\section{Related Work}
\label{sec:related-work}

\subsection{LLM-Based Multi-Agent Systems}

LLM-based multi-agent systems decompose complex tasks into interacting roles such as planners, critics, retrievers, coders, and answerers. Representative systems such as CAMEL, AutoGen, and ChatDev demonstrate that role-based conversation and workflow orchestration can improve task solving and software development~\citep{li2023camel,wu2023autogen,qian2024chatdev}. These systems establish the usefulness of multi-agent interaction, but they are primarily designed as inference-time orchestration frameworks: agents are usually distinguished by prompts, tools, and workflow positions, while the parameters underlying the agents are not jointly optimized as part of the multi-agent system. UnityMAS-O is complementary to this line of work: it treats such user-defined workflows as trainable objects and provides an optimization interface over the agents, their model mappings, and their rewards.

\subsection{Agentic RAG and Search with Multi-Agent Reinforcement Learning}

Retrieval-augmented generation and agentic search provide natural testbeds for multi-agent optimization because retrieval, filtering, reasoning, and answer generation can be assigned to different roles. MMOA-RAG formulates a multi-module RAG pipeline as a cooperative MARL problem and optimizes the modules toward final-answer quality~\citep{chen2025mmoarag}. MAO-ARAG introduces an adaptive RAG planner that selects executor agents for each query and trains the planner with answer-quality and cost rewards~\citep{chen2025maoarag}. M-ASK decomposes agentic search into search-behavior and knowledge-management roles with turn-level rewards~\citep{chen2026mask}. JADE further studies dynamic agentic RAG and jointly optimizes planning and execution under a shared backbone to address the mismatch between high-level strategies and local executors~\citep{chen2026jade}.

These works are closely related to our QA workflows and motivate the use of MARL for agentic search. However, they are usually designed around particular RAG or search architectures, often with fully shared parameters across functional roles. UnityMAS-O generalizes this direction by making the workflow graph, the agent-specific reward definitions, and the agent--model mapping explicit framework-level objects. This allows the same optimization substrate to instantiate shared, partially shared, or fully independent multi-agent systems beyond RAG.

\subsection{Reinforcement Learning for Multi-Agent LLM Systems}

Several recent works directly study reinforcement learning for LLM-based multi-agent systems. STRONGER-MAS proposes agent- and turn-wise grouped RL for collaborative LLMs and builds a system that supports MAS rollouts and updates for one or more policy models~\citep{zhao2026strongermas}. Dr. MAS analyzes instability in GRPO-style training for multi-agent LLM systems and proposes agent-wise advantage normalization to stabilize gradients~\citep{feng2026drmas}. MARTI provides a framework for multi-agent reinforced training and inference, combining centralized multi-agent interaction with distributed policy training and workflow-level reward allocation~\citep{zhang2026marti}. HACRL studies a different but related collaborative optimization setting in which heterogeneous agents share verified rollouts during training while executing independently at inference time~\citep{zhang2026hacrl}.

UnityMAS-O shares the goal of making LLM-based multi-agent systems trainable, but emphasizes a general abstraction that jointly represents logical agent roles, configurable agent--model mappings, graph-structured trajectories, and role-specific reward functions. More broadly, UnityMAS-O is designed as a general optimization interface for multi-agent systems defined by arbitrary tasks, arbitrary agent compositions, and user-defined reward functions.

\begin{table}[t]
\centering
\scriptsize
\setlength{\tabcolsep}{3pt}
\renewcommand{\arraystretch}{1.16}
\resizebox{\linewidth}{!}{
\begin{tabular}{p{2.1cm}p{3.2cm}p{3.2cm}p{3.2cm}p{3.6cm}}
\toprule
Aspect & MARTI~\citep{zhang2026marti} & Dr. MAS~\citep{feng2026drmas} & STRONGER-MAS~\citep{zhao2026strongermas} & UnityMAS-O \\
\midrule
Key distinction
& Scales MAS training and inference, but workflow graphs, sharing regimes, and reward attribution are less central as framework objects
& Stabilizes GRPO-style MAS training, but does not target general role-specific reward allocation as the main interface
& Repairs GRPO grouping with agent and turn identities, but remains organized around grouped-sample GRPO optimization
& Makes graph-structured workflows, explicit agent--model mappings, model-local data ownership, and PPO-style role rewards central to the framework \\
\midrule
Primary focus
& Scalable reinforced training and inference for LLM-based MAS
& Stable GRPO-style MAS training via agent-wise advantage normalization
& Agent- and turn-wise GRPO grouping for collaborative LLMs
& General optimization interface for user-defined LLM multi-agent workflows \\
\midrule
Reward and credit assignment
& Uses centralized reward models and reward shaping to decompose global rewards into agent-level rewards
& Mainly addresses stable normalization of GRPO advantages; reward assignment is still tied to group/outcome-style GRPO data
& Mixes global team reward with task-specific local rewards under AT-GRPO grouping
& First-class role-, turn-, and trajectory-level reward functions; delayed and delta rewards can be committed to the exact logical role and mapped model \\
\midrule
RL objective
& Supports several OpenRLHF algorithms, including REINFORCE++, GRPO, and PPO, with a shared training strategy across agents
& GRPO-style objective with agent-wise advantage normalization
& AT-GRPO; groups samples by agent and turn to repair standard GRPO grouping in MAS
& Multi-agent PPO-style training; does not require grouped candidates for reward normalization and naturally supports dense, sparse, delayed, and role-specific rewards \\
\midrule
Agent--model relation
& Distributed policy training for individual agents; sharing regimes are not the central framework abstraction
& Flexible agent-model assignment with optional sharing and heterogeneous model sizes
& Supports single-policy and multi-policy MAS training
& Explicit agent--model mapping \(\phi\), enabling full sharing, partial sharing, and full separation under one interface \\
\midrule
Resource and data path
& Multi-Agent World accesses vLLM engines and distributes rewarded trajectories to policy trainers
& Shared resource pooling and scheduling for actor backends, with trajectories split into per-model micro-batches
& Independent LLM resource pool per model plus CPU environment workers; routed experiences form per-model training batches
& Separate model-local worker groups for different LLMs; thin workflow metadata stays in the controller, while fat rollout tensors, log-probs, values, and buffers remain model-local for efficient training \\
\midrule
Workflow scope
& Built-in and custom dynamic workflows, evaluated mainly on mathematical reasoning
& Math reasoning and multi-turn search with heterogeneous agent-model assignments
& Game, planning, coding, and math workflows under AT-GRPO
& Retrieval-augmented QA, iterative agentic search, and reflective code generation; framework abstraction is graph-first rather than benchmark-first \\
\bottomrule
\end{tabular}}
\caption{Comparison with MARTI, Dr. MAS, and STRONGER-MAS. MARTI emphasizes scalable MAS reinforced training and inference, Dr. MAS focuses on stabilizing GRPO-style MAS optimization, and STRONGER-MAS adapts GRPO grouping to agent and turn identities. UnityMAS-O instead emphasizes a graph-first optimization interface with explicit agent--model mappings, model-local data ownership, and PPO-style role-specific reward assignment.}
\label{tab:marl-vs-unitymas}
\end{table}

Table~\ref{tab:marl-vs-unitymas} summarizes the positioning of UnityMAS-O relative to the most closely related LLM-MARL frameworks. MARTI establishes an important scalable training-and-inference stack for LLM-based multi-agent systems, but its main emphasis is system scalability rather than making workflow graphs, parameter-sharing regimes, and role-specific reward attribution first-class abstractions. Dr. MAS targets a different bottleneck: the instability of GRPO-style training in multi-agent settings, which it addresses through agent-wise advantage normalization. STRONGER-MAS also remains in the GRPO family, adapting grouped optimization to agent and turn identities through AT-GRPO.

UnityMAS-O differs from these systems in three design choices. First, rewards are assigned through a role-aware reward interface rather than mainly through global/group-level GRPO normalization or post-hoc reward shaping. This makes delayed rewards, marginal-improvement rewards, and heterogeneous role rewards easier to express in one workflow. Second, UnityMAS-O separates workflow-level control metadata from model-local rollout tensors, keeping heavy training data such as log probabilities, values, masks, and rollout buffers close to the model workers that consume it. Third, UnityMAS-O adopts PPO-style multi-agent optimization, making dense, sparse, delayed, and role-specific rewards easier to incorporate without constructing grouped candidate sets. These choices make UnityMAS-O complementary to GRPO-oriented MAS training methods: it is designed less as a single optimization recipe and more as a reusable substrate for graph-structured, role-aware multi-agent workflow optimization.

\subsection{LLM Post-Training and Distributed RL Infrastructure}

Modern LLM post-training frameworks have made RL-based optimization increasingly practical. TRL provides widely used interfaces for transformer-based reinforcement learning~\citep{vonwerra2020trl}; OpenRLHF focuses on scalable and efficient RLHF training~\citep{hu2024openrlhf}; slime builds an SGLang-native RL scaling stack~\citep{lmsys2025slime}; and verl/HybridFlow provides flexible RLHF execution with decoupled generation and training flows~\citep{sheng2024hybridflow}. These systems provide important infrastructure for rollout, optimization, and distributed execution, and UnityMAS-O builds on this ecosystem.

The key difference is the unit of abstraction. Existing post-training systems mainly optimize a single trainable policy model, even when the rollout environment includes tools or multi-turn interactions. UnityMAS-O lifts the optimization target from a single input--output trajectory to a graph-structured multi-agent trajectory. This requires workflow scheduling, agent-specific reward assembly, model-local buffering, and synchronized updates for one or more physical LLM instances.

\subsection{Distributed Execution and Model Serving}

Scalable multi-agent RL requires efficient distributed execution because workflow rollouts may contain many agent invocations, tool calls, and heterogeneous model requests. Ray provides a general distributed execution layer for remote task scheduling and resource management~\citep{moritz2018ray}, while vLLM improves LLM serving efficiency through PagedAttention~\citep{kwon2023vllm}. UnityMAS-O uses this systems perspective to separate workflow coordination from model-local training: a central controller manages graph execution and reward commitment, while worker groups handle generation, buffering, and PPO-style updates~\citep{schulman2017ppo}.

\paragraph{Positioning of UnityMAS-O.}
Prior LLM-MAS frameworks primarily provide orchestration, RAG-specific MARL methods optimize particular shared-parameter search workflows, and RL post-training frameworks primarily target single-model optimization. UnityMAS-O connects these directions by treating a user-defined multi-agent workflow as the unit of reinforcement learning. It exposes logical roles, agent--model mappings, graph-structured trajectories, role-specific rewards, and distributed model-local updates under one interface, enabling systematic optimization and study of LLM-based multi-agent systems.

\section{Conclusion}

This report presents \textbf{UnityMAS-O}, a general optimization framework for LLM-based multi-agent systems. The central idea is to treat a user-defined multi-agent workflow as the unit of reinforcement learning: logical roles, graph-structured trajectories, agent--model mappings, and role-specific rewards are represented explicitly, while physical LLM parameters can be fully shared, fully separated, or partially shared. Built on Ray and \texttt{verl}, UnityMAS-O provides a distributed runtime for workflow execution, reward assembly, asynchronous rollout, model-local buffering, advantage computation, and per-model policy updates.

The experiments show that this abstraction is useful in practice. On QA and agentic search workflows, MARL training improves validation F1 across datasets, workflows, and model scales, with particularly large gains for smaller backbones. On reflective code generation, the framework improves strict held-out all-passed rates and reduces the average number of verification turns needed before termination. The parameter-sharing study further shows that logical multi-agent workflows can still be trained when several roles share one physical LLM, supporting more flexible resource--performance trade-offs.

These results suggest that the main bottleneck in many LLM-based multi-agent systems is not only the design of a workflow, but the lack of a principled optimization interface for the workflow after it is designed. UnityMAS-O addresses this gap by converting manually specified multi-agent systems into trainable multi-agent policies. Beyond the experiments in this report, the framework provides infrastructure for studying collaboration, specialization, scaling behavior, heterogeneous resource allocation, and training stability in LLM-based multi-agent optimization.

\phantomsection
\section*{Contributions}
\label{sec:contributions}

\textbf{Authors.}
\textbf{Project lead and primary contributor:}
Yiqun Chen\textsuperscript{1}\textsuperscript{*}.
\textbf{Other contributors} are listed alphabetically by given name:
Bin Zhang\textsuperscript{2,3},
Biqing Qi\textsuperscript{4},
Erhan Zhang\textsuperscript{1},
Haitao Li\textsuperscript{5},
Jiaxin Mao\textsuperscript{1}\textsuperscript{\textdagger},
Jinyuan Feng\textsuperscript{2,3},
Lingyong Yan\textsuperscript{2},
Qi Liu\textsuperscript{1},
Rui Li\textsuperscript{1},
Shijie Wang\textsuperscript{4},
Wei Yang\textsuperscript{6},
Xiaochi Wei\textsuperscript{7},
Yan Gao\textsuperscript{7},
Yao Hu\textsuperscript{7},
Yi Wu\textsuperscript{7},
and Zechun Niu\textsuperscript{1}.
Detailed contributions are described below.

\textbf{Affiliations.}
\textsuperscript{1}Renmin University of China,
\textsuperscript{2}University of Chinese Academy of Sciences,
\textsuperscript{3}Institute of Automation, Chinese Academy of Sciences,
\textsuperscript{4}Shanghai AI Laboratory,
\textsuperscript{5}Tsinghua University,
\textsuperscript{6}University of Southern California,
\textsuperscript{7}Xiaohongshu Inc.
\textsuperscript{*}Project lead and primary contributor.
\textsuperscript{\textdagger}Corresponding author.

\textbf{Acknowledgement.}
We thank all contributors for their support and valuable suggestions. This work was conducted during Yiqun Chen's internship at Xiaohongshu Inc.

\subsection*{Detailed Contributions}

\textbf{Project leadership and core contributions.}
Yiqun Chen\textsuperscript{1}\textsuperscript{*} initiated and led the UnityMAS-O project, and served as the core contributor across all dimensions of the work, including framework design, system implementation, workflow formulation, experimental design, result analysis, and report writing.

\textbf{QA experiments.}
Erhan Zhang\textsuperscript{1} served as the primary collaborator for the QA experiments and assisted with retrieval-augmented QA and agentic search workflow evaluation. Qi Liu\textsuperscript{1}, Zechun Niu\textsuperscript{1}, Rui Li\textsuperscript{1}, and Jinyuan Feng\textsuperscript{2,3} also participated in the QA experiments.

\textbf{Code experiments.}
Shijie Wang\textsuperscript{4} and Biqing Qi\textsuperscript{4} assisted with the construction of the code workflow and the verifier used in the reflective code generation experiments.

\textbf{Math and embodied-agent experiments.}
Wei Yang\textsuperscript{6} assisted with experiments in math reasoning, embodied-agent environments, web-interaction tasks, and software-engineering tasks, including ALFWorld, WebShop, and SWE-bench. These experiments are part of ongoing work and are not included in this report.

\textbf{Report writing.}
Erhan Zhang\textsuperscript{1} provided assistance and feedback during report writing.

\textbf{Discussions.}
Wei Yang\textsuperscript{6}, Erhan Zhang\textsuperscript{1}, Haitao Li\textsuperscript{5}, Lingyong Yan\textsuperscript{2}, Bin Zhang\textsuperscript{2,3}, and others contributed valuable discussions on multi-agent optimization, workflow design, system implementation, and experimental analysis.

\textbf{Supervision and guidance.}
Jiaxin Mao\textsuperscript{1}\textsuperscript{\textdagger} is the corresponding author and provided supervision, high-level guidance, and important suggestions across the project. Xiaochi Wei\textsuperscript{7}, Yan Gao\textsuperscript{7}, Yi Wu\textsuperscript{7}, and Yao Hu\textsuperscript{7} provided valuable feedback from the perspectives of infrastructure design, system implementation, and optimization efficiency at Xiaohongshu Inc., where this work was conducted during Yiqun Chen\textsuperscript{1}\textsuperscript{*}'s internship.

\clearpage

\bibliography{colm2026_conference}

@inproceedings{Vaswani+2017,
 author = {Vaswani, Ashish and Shazeer, Noam and Parmar, Niki and Uszkoreit, Jakob and Jones, Llion and Gomez, Aidan N and Kaiser, \L ukasz and Polosukhin, Illia},
 booktitle = {Advances in Neural Information Processing Systems},
 pages = {},
 publisher = {Curran Associates, Inc.},
 title = {Attention is All you Need},
 url = {https://proceedings.neurips.cc/paper_files/paper/2017/file/3f5ee243547dee91fbd053c1c4a845aa-Paper.pdf},
 volume = {30},
 year = {2017}
}

@inproceedings{brown2020language,
    title = {Language Models are Few-Shot Learners},
    author = {Brown, Tom B. and Mann, Benjamin and Ryder, Nick and Subbiah, Melanie and Kaplan, Jared and Dhariwal, Prafulla and Neelakantan, Arvind and Shyam, Pranav and Sastry, Girish and Askell, Amanda and others},
    booktitle = {Advances in Neural Information Processing Systems},
    volume = {33},
    pages = {1877--1901},
    year = {2020},
    url = {https://proceedings.neurips.cc/paper/2020/hash/1457c0d6bfcb4967418bfb8ac142f64a-Abstract.html}
}

@inproceedings{li2023camel,
    title = {{CAMEL}: Communicative Agents for ``Mind'' Exploration of Large Language Model Society},
    author = {Li, Guohao and Hammoud, Hasan Abed Al Kader and Itani, Hani and Khizbullin, Dmitrii and Ghanem, Bernard},
    booktitle = {Advances in Neural Information Processing Systems},
    year = {2023},
    url = {https://arxiv.org/abs/2303.17760},
    doi = {10.48550/arXiv.2303.17760}
}

@article{wu2023autogen,
    title = {{AutoGen}: Enabling Next-Gen {LLM} Applications via Multi-Agent Conversation},
    author = {Wu, Qingyun and Bansal, Gagan and Zhang, Jieyu and Wu, Yiran and Li, Beibin and Zhu, Erkang and Jiang, Li and Zhang, Xiaoyun and Zhang, Shaokun and Liu, Jiale and Awadallah, Ahmed Hassan and White, Ryen W. and Burger, Doug and Wang, Chi},
    journal = {arXiv preprint arXiv:2308.08155},
    year = {2023},
    url = {https://arxiv.org/abs/2308.08155},
    doi = {10.48550/arXiv.2308.08155}
}

@inproceedings{qian2024chatdev,
    title = {{ChatDev}: Communicative Agents for Software Development},
    author = {Qian, Chen and Liu, Wei and Liu, Hongzhang and Chen, Nuo and Dang, Yufan and Li, Jiahao and Yang, Cheng and Chen, Weize and Su, Yusheng and Cong, Xin and Xu, Juyuan and Li, Dahai and Liu, Zhiyuan and Sun, Maosong},
    booktitle = {Proceedings of the 62nd Annual Meeting of the Association for Computational Linguistics},
    pages = {15174--15186},
    year = {2024},
    url = {https://aclanthology.org/2024.acl-long.810/}
}

@misc{vonwerra2020trl,
    title = {{TRL}: Transformer Reinforcement Learning},
    author = {von Werra, Leandro and Belkada, Younes and Tunstall, Lewis and Beeching, Edward and Thrush, Tristan and Lambert, Nathan and Huang, Shengyi and Rasul, Kashif and Gallouedec, Quentin},
    howpublished = {\url{https://github.com/huggingface/trl}},
    year = {2020}
}

@article{hu2024openrlhf,
    title = {{OpenRLHF}: An Easy-to-use, Scalable and High-performance {RLHF} Framework},
    author = {Hu, Jian and Wu, Xibin and Shen, Wei and Liu, Jason Klein and Zhu, Zilin and Wang, Weixun and Jiang, Songlin and Wang, Haoran and Chen, Hao and Chen, Bin and Fang, Weikai and Xianyu and Cao, Yu and Xu, Haotian and Liu, Yiming},
    journal = {arXiv preprint arXiv:2405.11143},
    year = {2024},
    url = {https://arxiv.org/abs/2405.11143},
    doi = {10.48550/arXiv.2405.11143}
}

@misc{lmsys2025slime,
    title = {slime: An {SGLang}-Native Post-Training Framework for {RL} Scaling},
    author = {{LMSYS Organization}},
    howpublished = {\url{https://lmsys.org/blog/2025-07-09-slime/}},
    year = {2025}
}

@article{sheng2024hybridflow,
    title = {{HybridFlow}: A Flexible and Efficient {RLHF} Framework},
    author = {Sheng, Guangming and Zhang, Chi and Ye, Zilingfeng and Wu, Xibin and Zhang, Wang and Zhang, Ru and Peng, Yanghua and Lin, Haibin and Wu, Chuan},
    journal = {arXiv preprint arXiv:2409.19256},
    year = {2024},
    url = {https://arxiv.org/abs/2409.19256},
    doi = {10.48550/arXiv.2409.19256}
}

@inproceedings{moritz2018ray,
    title = {Ray: A Distributed Framework for Emerging {AI} Applications},
    author = {Moritz, Philipp and Nishihara, Robert and Wang, Stephanie and Tumanov, Alexey and Liaw, Richard and Liang, Eric and Elibol, Melih and Yang, Zongheng and Paul, William and Jordan, Michael I. and Stoica, Ion},
    booktitle = {13th USENIX Symposium on Operating Systems Design and Implementation},
    pages = {561--577},
    year = {2018},
    url = {https://www.usenix.org/conference/osdi18/presentation/moritz}
}

@inproceedings{kwon2023vllm,
    title = {Efficient Memory Management for Large Language Model Serving with {PagedAttention}},
    author = {Kwon, Woosuk and Li, Zhuohan and Zhuang, Siyuan and Sheng, Ying and Zheng, Lianmin and Yu, Cody Hao and Gonzalez, Joseph E. and Zhang, Hao and Stoica, Ion},
    booktitle = {Proceedings of the ACM SIGOPS 29th Symposium on Operating Systems Principles},
    pages = {611--626},
    year = {2023},
    url = {https://arxiv.org/abs/2309.06180},
    doi = {10.1145/3600006.3613165}
}

@article{schulman2017ppo,
    title = {Proximal Policy Optimization Algorithms},
    author = {Schulman, John and Wolski, Filip and Dhariwal, Prafulla and Radford, Alec and Klimov, Oleg},
    journal = {arXiv preprint arXiv:1707.06347},
    year = {2017},
    url = {https://arxiv.org/abs/1707.06347},
    doi = {10.48550/arXiv.1707.06347}
}

@article{schulman2015gae,
    title = {High-Dimensional Continuous Control Using Generalized Advantage Estimation},
    author = {Schulman, John and Moritz, Philipp and Levine, Sergey and Jordan, Michael and Abbeel, Pieter},
    journal = {arXiv preprint arXiv:1506.02438},
    year = {2015},
    url = {https://arxiv.org/abs/1506.02438},
    doi = {10.48550/arXiv.1506.02438}
}

@article{kwiatkowski-etal-2019-natural,
    title = {Natural Questions: A Benchmark for Question Answering Research},
    author = {Kwiatkowski, Tom and Palomaki, Jennimaria and Redfield, Olivia and Collins, Michael and Parikh, Ankur and Alberti, Chris and Epstein, Danielle and Polosukhin, Illia and Devlin, Jacob and Lee, Kenton and Toutanova, Kristina and Jones, Llion and Kelcey, Matthew and Chang, Ming-Wei and Dai, Andrew M. and Uszkoreit, Jakob and Le, Quoc and Petrov, Slav},
    journal = {Transactions of the Association for Computational Linguistics},
    volume = {7},
    year = {2019},
    address = {Cambridge, MA},
    publisher = {MIT Press},
    url = {https://aclanthology.org/Q19-1026/},
    doi = {10.1162/tacl_a_00276},
    pages = {452--466}
}

@inproceedings{yang2018hotpotqa,
    title = {{HotpotQA}: A Dataset for Diverse, Explainable Multi-hop Question Answering},
    author = {Yang, Zhilin and Qi, Peng and Zhang, Saizheng and Bengio, Yoshua and Cohen, William W. and Salakhutdinov, Ruslan and Manning, Christopher D.},
    booktitle = {Conference on Empirical Methods in Natural Language Processing ({EMNLP})},
    year = {2018}
}

@article{jain2024livecodebench,
    title = {{LiveCodeBench}: Holistic and Contamination Free Evaluation of Large Language Models for Code},
    author = {Jain, Naman and Han, King and Gu, Alex and Li, Wen-Ding and Yan, Fanjia and Zhang, Tianjun and Wang, Sida and Solar-Lezama, Armando and Sen, Koushik and Stoica, Ion},
    journal = {arXiv preprint arXiv:2403.07974},
    year = {2024},
    url = {https://arxiv.org/abs/2403.07974},
    doi = {10.48550/arXiv.2403.07974}
}

@article{li2023taco,
    title = {{TACO}: Topics in Algorithmic {CO}de Generation Dataset},
    author = {Li, Rongao and Fu, Jie and Zhang, Bo-Wen and Huang, Tao and Sun, Zhihong and Lyu, Chen and Liu, Guang and Jin, Zhi and Li, Ge},
    journal = {arXiv preprint arXiv:2312.14852},
    year = {2023},
    url = {https://arxiv.org/abs/2312.14852},
    doi = {10.48550/arXiv.2312.14852}
}

@article{yang2025qwen3,
    title = {{Qwen3} Technical Report},
    author = {Yang, An and Li, Anfeng and Yang, Baosong and Zhang, Beichen and Hui, Binyuan and Zheng, Bo and Yu, Bowen and Gao, Chang and Huang, Chengen and Lv, Chenxu and Zheng, Chujie and Liu, Dayiheng and Zhou, Fan and Huang, Fei and Hu, Feng and Ge, Hao and Wei, Haoran and Lin, Huan and Tang, Jialong and Yang, Jian and others},
    journal = {arXiv preprint arXiv:2505.09388},
    year = {2025},
    url = {https://arxiv.org/abs/2505.09388},
    doi = {10.48550/arXiv.2505.09388}
}

@misc{deepcoder2025,
    title = {{DeepCoder}: A Fully Open-Source 14B Coder at O3-mini Level},
    author = {Luo, Michael and Tan, Sijun and Huang, Roy and Patel, Ameen and Ariyak, Alpay and Wu, Qingyang and Shi, Xiaoxiang and Xin, Rachel and Cai, Colin and Weber, Maurice and Zhang, Ce and Li, Li Erran and Popa, Raluca Ada and Stoica, Ion},
    howpublished = {\url{https://pretty-radio-b75.notion.site/DeepCoder-A-Fully-Open-Source-14B-Coder-at-O3-mini-Level-1cf81902c14680b3bee5eb349a512a51}},
    note = {Notion Blog},
    year = {2025}
}

@article{chen2026mask,
    title = {Beyond Monolithic Architectures: A Multi-Agent Search and Knowledge Optimization Framework for Agentic Search},
    author = {Chen, Yiqun and Yan, Lingyong and Yang, Zixuan and Zhang, Erhan and Zhao, Jiashu and Wang, Shuaiqiang and Yin, Dawei and Mao, Jiaxin},
    journal = {arXiv preprint arXiv:2601.04703},
    year = {2026},
    url = {https://arxiv.org/abs/2601.04703}
}

@inproceedings{chen2025mmoarag,
    title = {Improving Retrieval-Augmented Generation through Multi-Agent Reinforcement Learning},
    author = {Chen, Yiqun and Yan, Lingyong and Sun, Weiwei and Ma, Xinyu and Zhang, Yi and Wang, Shuaiqiang and Yin, Dawei and Yang, Yiming and Mao, Jiaxin},
    booktitle = {Advances in Neural Information Processing Systems},
    year = {2025},
    url = {https://arxiv.org/abs/2501.15228},
    doi = {10.48550/arXiv.2501.15228}
}

@article{chen2025maoarag,
    title = {{MAO-ARAG}: Multi-Agent Orchestration for Adaptive Retrieval-Augmented Generation},
    author = {Chen, Yiqun and Zhang, Erhan and Yan, Lingyong and Wang, Shuaiqiang and Huang, Jizhou and Yin, Dawei and Mao, Jiaxin},
    journal = {arXiv preprint arXiv:2508.01005},
    year = {2025},
    url = {https://arxiv.org/abs/2508.01005},
    doi = {10.48550/arXiv.2508.01005}
}

@article{chen2026jade,
    title = {{JADE}: Bridging the Strategic-Operational Gap in Dynamic Agentic {RAG}},
    author = {Chen, Yiqun and Zhang, Erhan and Hu, Tianyi and Wang, Shijie and Yang, Zixuan and Zhong, Meizhi and Wei, Xiaochi and Gao, Yan and Wu, Yi and Hu, Yao and Mao, Jiaxin},
    journal = {arXiv preprint arXiv:2601.21916},
    year = {2026},
    url = {https://arxiv.org/abs/2601.21916},
    doi = {10.48550/arXiv.2601.21916}
}

@inproceedings{zhao2026strongermas,
    title = {{STRONGER-MAS}: Multi-Agent Reinforcement Learning for Collaborative {LLM}s},
    author = {Zhao, Yujie and Hu, Lanxiang and Wang, Yang and Hou, Minmin and Zhang, Hao and Ding, Ke and Zhao, Jishen},
    booktitle = {The Fourteenth International Conference on Learning Representations},
    year = {2026},
    url = {https://openreview.net/forum?id=IdF6JqXWzx}
}

@article{feng2026drmas,
    title = {Dr. {MAS}: Stable Reinforcement Learning for Multi-Agent {LLM} Systems},
    author = {Feng, Lang and Zheng, Longtao and He, Shuo and Zhang, Fuxiang and An, Bo},
    journal = {arXiv preprint arXiv:2602.08847},
    year = {2026},
    url = {https://arxiv.org/abs/2602.08847},
    doi = {10.48550/arXiv.2602.08847}
}

@inproceedings{zhang2026marti,
    title = {{MARTI}: A Framework for Multi-Agent {LLM} Systems Reinforced Training and Inference},
    author = {Zhang, Kaiyan and Tian, Kai and Liu, Runze and Zeng, Sihang and Zhu, Xuekai and Jia, Guoli and Fan, Yuchen and Lv, Xingtai and Zuo, Yuxin and Jiang, Che and Wang, Yuru and Wang, Jianyu and Hua, Ermo and Long, Xinwei and Gao, Junqi and Sun, Youbang and Ma, Zhiyuan and Cui, Ganqu and Ding, Ning and Qi, Biqing and Zhou, Bowen},
    booktitle = {The Fourteenth International Conference on Learning Representations},
    year = {2026},
    url = {https://openreview.net/forum?id=E7jZqo0A50}
}

@article{zhang2026hacrl,
    title = {Heterogeneous Agent Collaborative Reinforcement Learning},
    author = {Zhang, Zhixia and Huang, Zixuan and Xia, Xin and Wang, Deqing and Zhuang, Fuzhen and Ma, Shuai and Ding, Ning and Yang, Yaodong and Li, Jianxin and Ban, Yikun},
    journal = {arXiv preprint arXiv:2603.02604},
    year = {2026},
    url = {https://arxiv.org/abs/2603.02604},
    doi = {10.48550/arXiv.2603.02604}
}

@inproceedings{shridhar2021alfworld,
    title = {{ALFWorld}: Aligning Text and Embodied Environments for Interactive Learning},
    author = {Shridhar, Mohit and Yuan, Xingdi and C{\^o}t{\'e}, Marc-Alexandre and Bisk, Yonatan and Trischler, Adam and Hausknecht, Matthew},
    booktitle = {International Conference on Learning Representations},
    year = {2021},
    url = {https://arxiv.org/abs/2010.03768}
}

@inproceedings{yao2022webshop,
    title = {{WebShop}: Towards Scalable Real-World Web Interaction with Grounded Language Agents},
    author = {Yao, Shunyu and Chen, Howard and Yang, John and Narasimhan, Karthik},
    booktitle = {Advances in Neural Information Processing Systems},
    year = {2022},
    url = {https://arxiv.org/abs/2207.01206}
}

@inproceedings{jimenez2024swebench,
    title = {{SWE}-bench: Can Language Models Resolve Real-World GitHub Issues?},
    author = {Jimenez, Carlos E. and Yang, John and Wettig, Alexander and Yao, Shunyu and Pei, Kexin and Press, Ofir and Narasimhan, Karthik},
    booktitle = {International Conference on Learning Representations},
    year = {2024},
    url = {https://arxiv.org/abs/2310.06770}
}
\bibliographystyle{colm2026_conference}


\clearpage

\section*{Appendix A: Full QA Training Gains}
\begin{table}[h]
\centering
\small
\setlength{\tabcolsep}{6pt}
\renewcommand{\arraystretch}{1.05}
\caption{Full QA validation gains across datasets, workflows, and model scales. We report validation F1 before multi-agent RL training and the best validation F1 achieved during training.}
\label{tab:qa-gains-full}
\begin{tabular}{llccccc}
\toprule
Dataset & Workflow & Size & Before RL & Best after RL & Abs. gain & Rel. gain \\
\midrule
\multirow{14}{*}{NQ}
& \multirow{5}{*}{QD-Answer} & 0.5B & 0.022 & 0.445 & +0.424 & +1943\% \\
& & 1.5B & 0.073 & 0.508 & +0.435 & +599\% \\
& & 3B & 0.266 & 0.535 & +0.269 & +101\% \\
& & 7B & 0.315 & 0.536 & +0.221 & +70\% \\
& & 14B & 0.328 & 0.594 & +0.267 & +81\% \\
\cmidrule(lr){2-7}
& \multirow{5}{*}{QD-Evidence-Answer} & 0.5B & 0.051 & 0.429 & +0.379 & +746\% \\
& & 1.5B & 0.070 & 0.480 & +0.410 & +588\% \\
& & 3B & 0.107 & 0.473 & +0.366 & +342\% \\
& & 7B & 0.279 & 0.509 & +0.231 & +83\% \\
& & 14B & 0.279 & 0.575 & +0.295 & +106\% \\
\cmidrule(lr){2-7}
& \multirow{4}{*}{M-ASK} & 0.5B & 0.015 & 0.142 & +0.127 & +842\% \\
& & 1.5B & 0.132 & 0.515 & +0.383 & +290\% \\
& & 3B & 0.259 & 0.513 & +0.253 & +98\% \\
& & 7B & 0.260 & 0.621 & +0.361 & +139\% \\
\midrule
\multirow{14}{*}{HotpotQA}
& \multirow{5}{*}{QD-Answer} & 0.5B & 0.032 & 0.397 & +0.366 & +1161\% \\
& & 1.5B & 0.069 & 0.493 & +0.424 & +612\% \\
& & 3B & 0.162 & 0.493 & +0.332 & +205\% \\
& & 7B & 0.284 & 0.523 & +0.240 & +85\% \\
& & 14B & 0.306 & 0.542 & +0.236 & +77\% \\
\cmidrule(lr){2-7}
& \multirow{5}{*}{QD-Evidence-Answer} & 0.5B & 0.068 & 0.359 & +0.291 & +430\% \\
& & 1.5B & 0.071 & 0.435 & +0.364 & +514\% \\
& & 3B & 0.107 & 0.473 & +0.366 & +342\% \\
& & 7B & 0.235 & 0.486 & +0.251 & +107\% \\
& & 14B & 0.279 & 0.521 & +0.243 & +87\% \\
\cmidrule(lr){2-7}
& \multirow{4}{*}{M-ASK} & 0.5B & 0.030 & 0.226 & +0.196 & +648\% \\
& & 1.5B & 0.127 & 0.525 & +0.398 & +313\% \\
& & 3B & 0.201 & 0.529 & +0.327 & +162\% \\
& & 7B & 0.254 & 0.573 & +0.319 & +126\% \\
\bottomrule
\end{tabular}
\end{table}

\end{document}